\documentclass[10pt,twocolumn,letterpaper]{article}

\usepackage{cvpr}
\usepackage{times}
\usepackage{epsfig}
\usepackage{graphicx}
\usepackage{amsmath}
\usepackage{amssymb}
\usepackage{epstopdf}
\usepackage{tabularx}
\usepackage{enumitem}
\usepackage{paralist}
\usepackage[pagebackref=true,breaklinks=true,letterpaper=true,colorlinks,bookmarks=false]{hyperref}

\cvprfinalcopy 

\def\cvprPaperID{7} 

\begin{document}

\title{MA$\boldsymbol{^3}$: Model Agnostic Adversarial Augmentation for Few Shot learning}

\author{
Rohit Jena
\qquad
Shirsendu Sukanta Halder
\qquad
Katia Sycara \\
{\tt\small rjena@cs.cmu.edu} \qquad
{\tt\small shirsenh@cs.cmu.edu} \qquad
{\tt\small katia@cs.cmu.edu} \\ \\
Carnegie Mellon University \\
}

\maketitle

\begin{abstract}
    Despite the recent developments in vision-related problems using deep neural networks, there still remains a wide scope in the improvement of generalizing these models to unseen examples. In this paper, we explore the domain of few-shot learning with a novel augmentation technique. In contrast to other generative augmentation techniques, where the distribution over input images are learnt, we propose to learn the probability distribution over the image transformation parameters which are easier and quicker to learn. Our technique is fully differentiable which enables its extension to versatile data-sets and base models. We evaluate our proposed method on multiple base-networks and 2 data-sets to establish the robustness and efficiency of this method. We obtain an improvement of nearly 4\% by adding our augmentation module without making any change in network architectures. We also make the code \footnote{https://github.com/rohitrango/STNAdversarial}  readily available for usage by the community.
\end{abstract}

\section{Introduction}
Supervised learning algorithms have demonstrated tremendous success in a multitude of tasks both high-level like classification \cite{simonyan2014very}, detection \cite{ren2015faster}, etc and also in low-level tasks such as segmentation \cite{long2015fully} after the explosion of deep neural networks. However, the same statement cannot be made for situations where the model is expected to generalize in the absence of densely available labels. This is unlike humans, who generalise in an incremental manner to novel classes by observing only a few number of examples \cite{lake2015human}. The importance of a learning model that improves on unseen examples on gathering more experience is instrumental in almost all practical problems where annotating labels is either not scalable or unavailable due to safety or privacy issues.

Motivated by the aforementioned issues, recent approaches to generalize learning models range from weakly-supervised learning \cite{bilen2016weakly}, transfer learning \cite{tan2018survey}, domain adaptation techniques \cite{patel2015visual}, data augmentation \cite{shorten2019survey}, incremental learning \cite{ren2019incremental} and task based few shot learning \cite{matchingnets, sun2019meta, protonets, sung2018learning}. Few-shot classification aims to accommodate to novel classes unseen during training by just using a few examples during test time. This is unlike fine-tuning, where the classifier uses a previously learnt representation and tunes its parameters to maximize accuracy over the new data. The problem with fine tuning is that the classifier would most likely overfit to the new data when it is given as few as five examples. In this work, we take inspiration from humans in the sense that in order for registration, we infer the scene from different perspectives and then are able to generalize in similar future settings. We present a novel method for end-to-end differentiable data augmentation technique inspired by Spatial Transformer Networks \cite{jaderberg2015spatial} and inference technique for single and few-shot learning scenarios.
Our contributions are as follows:
\begin{compactenum}
    \item We propose a theory for a new data-augmentation technique inspired from projective transformations in the 3D camera pinhole model.
    \item We demonstrate an algorithm that estimates the data augmentation parameters in an end-to-end neural network model to generalize under a multi-class \textit{k}-shot classification framework.
    \item We present analysis of our proposed algorithm using 3 recent few-shot learning paradigms and establish the efficiency of our method for one-shot and few-shot learning on two versatile datasets.
\end{compactenum}
The rest of the paper is as follows. Section \ref{sec:relwork} presents some of the previous works in literature pertaining to learning with limited labels and data augmentation techniques. Section \ref{sec:method} describes our method in detail followed by Section \ref{sec:exp} which shows detailed analysis and comparison. The final Section \ref{sec:conc} contains concluding remarks and discussions about scope for future work.

\section{Related Work}
\label{sec:relwork}

\textbf{Few-shot learning:}   Lake \etal \cite{lake2011one} propose a generative model and infer handwritten characters from latent strokes in new characters.
Ravi and Larochelle \cite{ravi2016optimization} use a LSTM-based \textit{meta-learner} that captures short-term knowledge particular to a task and long-term knowledge common to all tasks.
ProtoNets \cite{protonets} learn a representation based metric space and perform classification using the "prototypes" (class means) of each class. Vinyals \etal \cite{matchingnets} propose a network called Matching Networks that learns the mapping between a small labelled support set and an unlabelled example. The principle that the testing and training conditions should match is used for the training procedure. Few-shot learning has also been explored in the context of meta-learning by Finn \etal \cite{maml} where they propose an algorithm for fast adaptation of networks on versatile tasks and demonstrate their effectiveness on one-shot learning tasks. Finn \etal \cite{finn2017one} further explore the task of one-shot learning for a robot under the framework of meta-learning combined with imitation learning from visual demonstrations. Meta learning and transfer learning was combined by \cite{sun2019meta} to propose an efficient learning curriculum which they name \textit{hard-task meta batch scheme} that improves the convergence and accuracy.

\textbf{Data augmentation:} Antoniou \etal \cite{antoniou2017data} were the first to demonstrate improved performances on meta-learning tasks using data augmentation techniques. They do so by generalizing the model to generate class-agnostic data samples. Zhang \etal \cite{zhang2018metagan} approach the problem of few-shot learning using a unified adversarial generator that is capable of learning sharper boundaries for supervised few-shot and semi-supervised few-shot scenarios as well. This is facilitated by making the GAN generate fake data that provides additional examples for training. Our method is also based on adversarial training but instead of directly generating augmented examples for training, we generate the parameters for transforming the input to learn a robust classifier. The closest work compared to ours is \cite{cubuk2019autoaugment} where they use a search algorithm to search the best policy for augmenting a single sample in a mini-batch. The policies consist of sub-policies consisting of either rotation, translation or shearing functions.
However, the method is not tested in few-shot settings and the use of reinforcement learning can be unstable with an evolving reward function.
Our work is different in the sense that instead of considering these image processing functions independently, we use an adversarial scheme to learn the complete affine transform matrix elements which provides us with better generalization. We also show that a variant which predicts the parameters independently doesn't perform as well as our method.

\section{Method}
\label{sec:method}
Our model takes inspiration from how humans observe novel objects - they don't just register one ``snapshot'' of the object, but rather take a look from multiple coherent perspectives.
Although this may not be possible given that we do not have images of the same object taken from different perspectives, we can approximate it by assuming that the object is placed far away from the camera (i.e. $z \approx z_0 \gg 1$).

\begin{figure*}[!htb]
    \centering
    \includegraphics[width=0.75\textwidth]{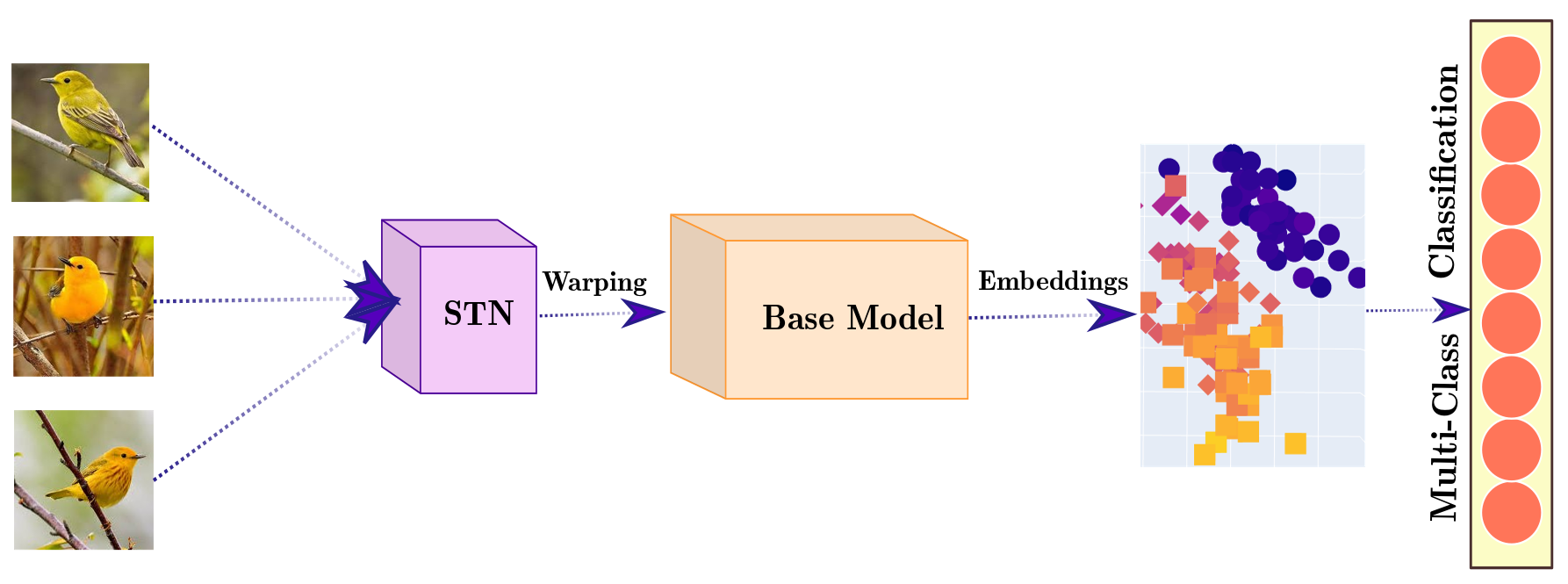}
    \caption{Proposed augmentation module to few shot learning}
    \label{fig:my_label}
\end{figure*}

Consider a 3D point of an object in homogeneous coordinates $\begin{pmatrix}x & y & z & 1\end{pmatrix}^T$ and its 2D projection into the image plane $\begin{pmatrix}u & v & 1\end{pmatrix}^T$.
Without loss of generality, assume that $R = I$, $t = 0$ to get $u_1 = x/z_0, v_1 = y/z_0$. \\
Consider a slight change of roll ($\gamma$), yaw ($\alpha$) and pitch ($\beta$) where $\|\gamma\|, \|\alpha\|, \|\beta\| \ll 1$, and a small change in translation $\boldsymbol{t}$ such that  $\| \boldsymbol{t}\| \ll 1$. Plugging these formulae into the rotation matrix and using Taylor expansion (ignoring third order terms and higher), we have:
\[
R = \begin{bmatrix} 1 - \frac{\alpha^2}{2} - \frac{\beta^2}{2}  & \beta\gamma - \alpha  & \beta + \alpha\gamma \\
                    \alpha  & 1 - \frac{\alpha^2}{2} - \frac{\gamma^2}{2}  & \alpha\beta - \gamma \\
                    -\beta  & \gamma  & 1 - \frac{\beta^2}{2} - \frac{\gamma^2}{2}
\end{bmatrix}
\]
and \[
t = \begin{bmatrix} t_x && t_y && t_z \end{bmatrix}^T
\]
The new point in the image plane corresponding to the original 3D coordinate is:
\[
    u_2 = \frac{(1 - \frac{\alpha^2}{2} - \frac{\beta^2}{2})x + (\beta\gamma - \alpha)y + (\beta + \alpha\gamma)z_0 + t_x}{-\beta x + \gamma y + (1 - \frac{\beta^2}{2} - \frac{\gamma^2}{2})z_0 + t_z}
\]
Since we assume it to be a distant object, and the values of $\alpha, \beta, \gamma$ are relatively small, the denominator can be simplified using binomial expansion
$$\frac{1}{z_0(1 - \delta)}\approx \frac{1 + \delta}{z_0}$$
where $\delta = \frac{\beta^2}{2} + \frac{\gamma^2}{2} + \frac{\beta x}{z_0} - \frac{\gamma y}{z_0} + \frac{t_z}{z_0}$
The new point on the image plane is approximated as

\begin{equation*}
\resizebox{0.49\textwidth}{!}{
 $
     u_2 \approx (1 + \delta)\Bigg[ \left( 1 - \frac{\alpha^2}{2} - \frac{\beta^2}{2} \right) \frac{x}{z_0} + \left( \beta\gamma - \alpha \right) \frac{y}{z_0} + \left( \beta + \alpha\gamma + \frac{t_x}{z_0} \right) \Bigg]
 $
}
\end{equation*}
and
\begin{equation*}
\resizebox{0.49\textwidth}{!}{
$
    v_2 \approx (1 + \delta) \Bigg[ \alpha \frac{x}{z_0} + \left( 1 - \frac{\gamma^2}{2} - \frac{\beta^2}{2} \right) \frac{y}{z_0} + \left( \alpha\beta - \gamma + \frac{t_y}{z_0} \right) \Bigg]
$
}
\end{equation*}
Substituting the values of $u_1, v_1$ we get
\begin{equation*}
    \resizebox{0.3\textwidth}{!}{
    $
     \begin{bmatrix} u_2 \\ v_2 \end{bmatrix}  = \begin{bmatrix} 1 + \delta_1  & \delta_2  & \delta_3 \\ \delta_4  & 1 + \delta_5  & \delta_6 \end{bmatrix} \begin{bmatrix} u_1 \\ v_1 \\ 1 \end{bmatrix}
     $
    }
\end{equation*}
where $\|\delta_i\| \ll 1, \forall i \in \{1 .. 6\} $
We approximate the distortion in rotation and translation using an affine transform of the given form, which encourages only slight deviation from the identity transform.
The values of the parameters $\delta_i$ can be determined using an adversary that detects the distortions that the model hasn't generalized to. This is the core idea which forms the basis of generalization to unseen examples. We use Spatial Transformer Networks (STN) which are end-to-end differentiable spatial manipulators.
STN computes parameters of the spatial manipulation rather than the manipulated image itself, making it easier to learn a few parameters and perform powerful spatial transformations.
They are generally used as a starting module to output a canonical version of an image that can be used as input to a classifier.
However, we use it in an adversarial manner by backpropagating through the Cross Entropy loss of the few-shot learner.
Learning the trend of the parameters $\delta_i$ is simpler and quicker than GANs that learn the data distribution over entire images in response to a noise signal or other support images.
We show that this form of augmentation to an image is more effective than applying standard augmentations like random rotations, translations and scaling.
At every epoch, the few shot learner processes a batch of support and query examples.
The few shot network minimizes the classification loss on the query examples given the support examples.
The Transformer takes gradients with respect to the support images to maximize the classification loss on the query images.
Let the transformer be a function $f$ parameterized by $\phi$ and the few shot learner is a function $g$ parameterized by $\theta$.
Let $S = \{s_1, s_2, \ldots s_n\}$ be the support dataset and $Q = \{q_1, q_2 \ldots q_m \}$ be the query dataset.
The optimization problem becomes:
$$ \max_\phi\min_\theta \sum_{i=1}^{m}L(g_\theta(q_i | f_\phi(s_1), f_\phi(s_2) \ldots f_\phi(s_n))) $$

To make sure that the Transformer doesn't deviate from the identity transform, we apply a regularization term that penalizes deviation from the identity affine transform. The regularization is given by the following term:
$$ L_{reg}(f_\phi(s)) = \left\| \begin{bmatrix} a_1(s) & a_2(s) & a_3(s) \\ a_4(s) & a_5(s) & a_6(s) \\ \end{bmatrix} - \begin{bmatrix} 1 & 0 & 0 \\ 0 & 1 & 0 \\ \end{bmatrix} \right\|^2$$
The modified optimization problem becomes:
\begin{align*}
    \max_\phi\min_\theta \sum_{i=1}^{m}L(g_\theta(q_i | f_\phi(s_1), f_\phi(s_2) \ldots f_\phi(s_n))) \\
    - \lambda \sum_{j=1}^{n}L_{reg}(f_\phi(s_j))
\end{align*}
where $\lambda$ is a hyperparameter.
Note that regularization plays an important role, because without any regularization the STN can morph the images to have unrecognizable features and hence maximizing the classification loss and not allowing the classifier to learn useful features.
Without explicit regularization, the parameters of the affine matrix predicted by the STN will also violate the assumption about the magnitudes of the $\delta$ parameters.
This does occur in our experiments when we set $\lambda = 0$, the accuracy over the validation set decreases because the classifier failed to learn good features during training.

\section{Experiments}
\label{sec:exp}
To analyse the effect of adversarial Spatial Transformer Networks, we test our training framework on the Omniglot \cite{omniglot} and MiniImageNet \cite{miniimagenet} datasets.
We show that our method is base-model agnostic by testing on 3 different methods - Prototypical Networks \cite{protonets}, Matching Networks \cite{matchingnets} and Model-Agnostic Meta Learning (MAML) \cite{maml} frameworks for few shot learning.
We observe that all baselines have very high accuracy on the Omniglot dataset, and adding an STN improves the results only marginally.
Therefore, we show results for Omniglot only with Prototypical Networks.
However, the improvements in accuracy for MiniImageNet are significant and we test our module with all the three baselines.

Prototypical networks received some concerns about reproducibility in results \cite{rmliclr}, \cite{issueprotonets1}, \cite{issueprotonets2}.
To provide consistent results for all methods, we use the code provided by \cite{fewshotgh} and incorporate our module into the code.

In standard classification tasks, the training data is augmented and the validation data is not augmented.
We follow the same procedure, we augment the meta-train (or support) examples and do not augment the meta-validation (or query) examples while training.
During test time, the STN is disabled for both support and query examples.
To avoid potential data distribution shift between the support examples encountered during the training phase and validation phase, we apply a dropout on the output of the STN to retain some of the support images (by randomly selecting images and setting their affine matrix to identity).
The dropout value is fixed to $0.5$ and the values of $\lambda$ are obtained using a coarse grid search on a log-scale and a finer grid search on a linear scale after choosing the best interval from the coarse search.
The first baseline does not use any data augmentation.
The second baseline uses standard data augmentation like random scaling, translation, and rotation.
However, unlike random data augmentation, our method outputs parameters by an adversarial STN.
The STN outputs the values of rotation $\theta$, translation $p_x, p_y$ and scale $s$ and the affine matrix is constructed as:
$$ A = \begin{bmatrix}s \cos(\theta) & -s \sin(\theta) & p_x  \\  s \sin(\theta) & s \cos(\theta) & p_y\end{bmatrix}$$
The values are bounded to $\theta \in [-\theta_0, \theta_0]$, $s \in [1 - \epsilon_s, 1 + \epsilon_s]$ and $p_x, p_y \in [-T, T]$ using tanh activations and appropriate scaling.
For all experiments, we set $\theta_0 = \pi$, $\epsilon_s = 0.1$, and $T = 0.1 \max(H, W)$, where $H, W$ are the height and width of the images.

\begin{table}[!htb]
\centering
\caption{Quantitative comparison of our method with baseline methods on Omnigot \cite{omniglot} dataset. The base network used in this scenario is ProtoNets \cite{protonets} with $h_{dim} = 128$ and $\gamma = 0.5$. The comparisons provided in both the tables are with vanilla-baseline method, baseline method with commonly used augmentation techniques, our proposed method with no constraint/regularization on the transformation parameters and our method with constrained parameters.}
\vspace{1em}
\label{tab:omniglot}
\resizebox{\linewidth}{!}{%
\begin{tabular}{c|cccc}
\textit{Classification Task} & \textit{Baseline} & \textit{\begin{tabular}[c]{@{}c@{}}Baseline\end{tabular}} &
\textit{\begin{tabular}[c]{@{}c@{}}Ours\end{tabular}} &
\textit{Ours} \\
& & (with standard aug.) & ($\lambda = 0$)& \\
\hline
20 way, 5 shot & 98.70\% & \textbf{98.89\%} & 94.25\% & 98.80\%          \\
20 way, 1 shot & 95.9\%  & \textbf{96.09\%} & 80.70\% & 95.97\%          \\
5 way, 5 shot  & 99.62\% & 99.62\%          & 99.40\%& \textbf{99.67\%} \\
5 way, 1 shot  & 98.42\% & 98.60\%          & 96.40\% & \textbf{98.61\%}
\end{tabular}
}
\end{table}

\begin{table}[!htb]
\centering
\caption{Quantitative Comparison of our method with baseline methods on miniImageNet \cite{miniimagenet} dataset. The first table contains results using ProtoNets, followed by MAML \cite{maml} followed by Matching Nets \cite{matchingnets} as the base network.}
\vspace{1em}
\label{tab:miniimage}
\resizebox{\linewidth}{!}{%
\begin{tabular}{c|cccc}
\textit{Classification Task} & \textit{Baseline} & \textit{\begin{tabular}[c]{@{}c@{}}Baseline\end{tabular}} &
\textit{\begin{tabular}[c]{@{}c@{}}Ours\end{tabular}} &
\textit{Ours} \\
(ProtoNets \cite{protonets})& & (with standard aug.) & $(\lambda = 0)$ & \\
\hline
5 way, 5 shot  & 66.6\% & 70.2\%  &  58.8\%      & \textbf{70.4\%} \\
5 way, 1 shot  & 51.4\% & 49.8\%  &  36.2\%  & \textbf{52.8\%}
\end{tabular}
}
\vspace{1em}

\resizebox{\linewidth}{!}{%
\begin{tabular}{c|cccc}
\textit{Classification Task} & \textit{Baseline} & \textit{\begin{tabular}[c]{@{}c@{}}Baseline\end{tabular}} &
\textit{\begin{tabular}[c]{@{}c@{}}Ours\end{tabular}} &
\textit{Ours} \\
(MAML \cite{maml})& & (with standard aug.) & $(\lambda = 0)$&\\
\hline
5 way, 5 shot  & 65.9\% & 66.3\% &    57.9\%  & \textbf{67.0\%} \\
5 way, 1 shot  & 47.3\% & 47.3\% &    32.1\%  & \textbf{48.2\%}
\end{tabular}
}
\vspace{1em}

\resizebox{\linewidth}{!}{%
\begin{tabular}{c|cccc}
\textit{Classification Task} & \textit{Baseline} & \textit{\begin{tabular}[c]{@{}c@{}}Baseline\end{tabular}} &
\textit{\begin{tabular}[c]{@{}c@{}}Ours\end{tabular}} &
\textit{Ours} \\
(Matching Nets \cite{matchingnets})& & (with standard aug.) & $(\lambda = 0)$ & \\
\hline
5 way, 5 shot  & 59.8\% & 61.4\%   &  47.8\%   & \textbf{62.0\%} \\
5 way, 1 shot  & 47.0\% & 48.4\%   &  34.2\%    & \textbf{50.8\%}
\end{tabular}
}
\end{table}
The improvements on Prototypical Networks for Omniglot dataset (Table \ref{tab:omniglot}) are not very significant because the baselines already learn features which are general enough to perform well on this easy dataset.
However, miniImageNet is a dataset with more variance and would require a classifier to learn complex features to perform well.
Our method bumps the performance of the base classifiers by as much as $3.8\%$ without requiring any change to the model architecture, thereby learning better features than that are learnt without the adversarial augmentation (Table \ref{tab:miniimage}).
Expectedly, our method fails to generalize in the absence of regularization as the STN exploits the freedom of choosing the affine matrix by performing transformations which produce images that are very far from the original data distribution and are often degenerate (for example, excessively zoomed images can result in the image being just a single color).
These images hinder the actual learning of the classifier and the accuracy drops significantly below the baseline method. This clearly reinforces our hypothesis regarding the importance of regularization while estimating the transformation parameters. Baseline with standard data augmentation performs better than the baseline in most cases, but the improvement is not consistent (see table \ref{tab:miniimage} - 5 way, 5 shot in ProtoNets and 5 way, 1 shot in MAML).

\section{Conclusion}
\label{sec:conc}
In this paper, we introduced MA$^3$, a model-agnostic adversarial augmentation technique for few shot learning.
The method is inspired by an approximate model of how humans ``cheat'' by observing a novel object from various perspectives.
We show that the model can be approximated using an affine transform, and Spatial Transformer Networks naturally fit into the equation by predicting affine transforms that the classifier is not robust to.
Experiments show that the method works on both metric-based and meta-learning approaches by testing it on top of 3 popularly known works - Prototypical Networks, Matching Networks and the MAML framework.
Our method performs better than standard augmentations, which raises the question as to which augmentations are actually useful in learning robust features, which is an interesting avenue for future work.

\clearpage
{\small
\bibliographystyle{ieee_fullname}
\bibliography{egbib}

\begin{thebibliography}{10}\itemsep=-1pt

\bibitem{issueprotonets2}
alirezazareian.
\newblock {Issue \#5: Reproducing Mini-Imagenet Results}.
\newblock \url{https://github.com/jakesnell/prototypical-networks/issues/5},
  2018.

\bibitem{antoniou2017data}
Antreas Antoniou, Amos Storkey, and Harrison Edwards.
\newblock Data augmentation generative adversarial networks.
\newblock {\em arXiv preprint arXiv:1711.04340}, 2017.

\bibitem{bilen2016weakly}
Hakan Bilen and Andrea Vedaldi.
\newblock Weakly supervised deep detection networks.
\newblock In {\em Proceedings of the IEEE Conference on Computer Vision and
  Pattern Recognition}, pages 2846--2854, 2016.

\bibitem{rmliclr}
Thomas Boquet, Laure Delisle, Denis Kochetkov, Nathan Schucher, Boris~N.
  Oreshkin, and Julien Cornebise.
\newblock Reproducibility and stability analysis in metric-based few-shot
  learning.
\newblock In {\em RML@ICLR}, 2019.

\bibitem{cubuk2019autoaugment}
Ekin~D Cubuk, Barret Zoph, Dandelion Mane, Vijay Vasudevan, and Quoc~V Le.
\newblock Autoaugment: Learning augmentation strategies from data.
\newblock In {\em Proceedings of the IEEE conference on computer vision and
  pattern recognition}, pages 113--123, 2019.

\bibitem{maml}
Chelsea Finn, Pieter Abbeel, and Sergey Levine.
\newblock Model-agnostic meta-learning for fast adaptation of deep networks.
\newblock In {\em Proceedings of the 34th International Conference on Machine
  Learning-Volume 70}, pages 1126--1135. JMLR. org, 2017.

\bibitem{finn2017one}
Chelsea Finn, Tianhe Yu, Tianhao Zhang, Pieter Abbeel, and Sergey Levine.
\newblock One-shot visual imitation learning via meta-learning.
\newblock {\em arXiv preprint arXiv:1709.04905}, 2017.

\bibitem{jaderberg2015spatial}
Max Jaderberg, Karen Simonyan, Andrew Zisserman, et~al.
\newblock Spatial transformer networks.
\newblock In {\em Advances in neural information processing systems}, pages
  2017--2025, 2015.

\bibitem{fewshotgh}
Oscar Knagg.
\newblock {Repository for few-shot learning machine learning projects}.
\newblock \url{https://github.com/oscarknagg/few-shot/}, 2018.

\bibitem{lake2011one}
Brenden Lake, Ruslan Salakhutdinov, Jason Gross, and Joshua Tenenbaum.
\newblock One shot learning of simple visual concepts.
\newblock In {\em Proceedings of the annual meeting of the cognitive science
  society}, volume~33, 2011.

\bibitem{omniglot}
Brenden Lake, Ruslan Salakhutdinov, Jason Gross, and Joshua Tenenbaum.
\newblock One shot learning of simple visual concepts.
\newblock In {\em Proceedings of the annual meeting of the cognitive science
  society}, 2011.

\bibitem{lake2015human}
Brenden~M Lake, Ruslan Salakhutdinov, and Joshua~B Tenenbaum.
\newblock Human-level concept learning through probabilistic program induction.
\newblock {\em Science}, 350(6266):1332--1338, 2015.

\bibitem{issueprotonets1}
Yanbin Liu.
\newblock {Issue \#2: Can you release detailed configuration?}
\newblock \url{https://github.com/jakesnell/prototypical-networks/issues/2},
  2018.

\bibitem{long2015fully}
Jonathan Long, Evan Shelhamer, and Trevor Darrell.
\newblock Fully convolutional networks for semantic segmentation.
\newblock In {\em Proceedings of the IEEE conference on computer vision and
  pattern recognition}, pages 3431--3440, 2015.

\bibitem{patel2015visual}
Vishal~M Patel, Raghuraman Gopalan, Ruonan Li, and Rama Chellappa.
\newblock Visual domain adaptation: A survey of recent advances.
\newblock {\em IEEE signal processing magazine}, 32(3):53--69, 2015.

\bibitem{ravi2016optimization}
Sachin Ravi and Hugo Larochelle.
\newblock Optimization as a model for few-shot learning.
\newblock 2016.

\bibitem{ren2019incremental}
Mengye Ren, Renjie Liao, Ethan Fetaya, and Richard Zemel.
\newblock Incremental few-shot learning with attention attractor networks.
\newblock In {\em Advances in Neural Information Processing Systems}, pages
  5276--5286, 2019.

\bibitem{ren2015faster}
Shaoqing Ren, Kaiming He, Ross Girshick, and Jian Sun.
\newblock Faster r-cnn: Towards real-time object detection with region proposal
  networks.
\newblock In {\em Advances in neural information processing systems}, pages
  91--99, 2015.

\bibitem{miniimagenet}
Olga Russakovsky, Jia Deng, Hao Su, Jonathan Krause, Sanjeev Satheesh, Sean Ma,
  Zhiheng Huang, Andrej Karpathy, Aditya Khosla, Michael Bernstein, et~al.
\newblock Imagenet large scale visual recognition challenge.
\newblock {\em International journal of computer vision}, 115(3):211--252,
  2015.

\bibitem{shorten2019survey}
Connor Shorten and Taghi~M Khoshgoftaar.
\newblock A survey on image data augmentation for deep learning.
\newblock {\em Journal of Big Data}, 6(1):60, 2019.

\bibitem{simonyan2014very}
Karen Simonyan and Andrew Zisserman.
\newblock Very deep convolutional networks for large-scale image recognition.
\newblock {\em arXiv preprint arXiv:1409.1556}, 2014.

\bibitem{protonets}
Jake Snell, Kevin Swersky, and Richard Zemel.
\newblock Prototypical networks for few-shot learning.
\newblock In {\em Advances in neural information processing systems}, pages
  4077--4087, 2017.

\bibitem{sun2019meta}
Qianru Sun, Yaoyao Liu, Tat-Seng Chua, and Bernt Schiele.
\newblock Meta-transfer learning for few-shot learning.
\newblock In {\em Proceedings of the IEEE Conference on Computer Vision and
  Pattern Recognition}, pages 403--412, 2019.

\bibitem{sung2018learning}
Flood Sung, Yongxin Yang, Li Zhang, Tao Xiang, Philip~HS Torr, and Timothy~M
  Hospedales.
\newblock Learning to compare: Relation network for few-shot learning.
\newblock In {\em Proceedings of the IEEE Conference on Computer Vision and
  Pattern Recognition}, pages 1199--1208, 2018.

\bibitem{tan2018survey}
Chuanqi Tan, Fuchun Sun, Tao Kong, Wenchang Zhang, Chao Yang, and Chunfang Liu.
\newblock A survey on deep transfer learning.
\newblock In {\em International conference on artificial neural networks},
  pages 270--279. Springer, 2018.

\bibitem{matchingnets}
Oriol Vinyals, Charles Blundell, Timothy Lillicrap, Daan Wierstra, et~al.
\newblock Matching networks for one shot learning.
\newblock In {\em Advances in neural information processing systems}, pages
  3630--3638, 2016.

\bibitem{zhang2018metagan}
Ruixiang Zhang, Tong Che, Zoubin Ghahramani, Yoshua Bengio, and Yangqiu Song.
\newblock Metagan: An adversarial approach to few-shot learning.
\newblock In {\em Advances in Neural Information Processing Systems}, pages
  2365--2374, 2018.

\end{thebibliography}
}

\end{document}